\documentclass[letterpaper, 10 pt, conference]{ieeeconf}  

\usepackage{colortbl}
\usepackage{pifont}
\usepackage[table,xcdraw]{xcolor}
\usepackage[bookmarks=true]{hyperref}
\usepackage{booktabs}
\definecolor{darkyellow}{rgb}{0.72, 0.53, 0.04}
\usepackage{multicol}
\usepackage{graphicx}
\newcommand{\yes}{\color{blue}{\ding{51}}}
\newcommand{\no}{\color{red}{\ding{55}}}
\usepackage{dingbat}
\usepackage{contour}
\usepackage{amsmath}
\usepackage{amssymb}
\IEEEoverridecommandlockouts                              

\title{\LARGE \bf
RwoR: Generating Robot Demonstrations from Human Hand \\Collection for Policy
Learning without Robot
}

\author{Liang Heng$^{1,2*}$, Xiaoqi Li$^{1,2*}$, Shangqing Mao$^{1,2}$, Jiaming Liu$^{3}$, Ruolin Liu$^{3}$, Jingli Wei$^{3}$, Yu-Kai Wang$^{3}$, \\Yueru Jia$^{3}$, Chenyang Gu$^{3}$, Rui Zhao$^{4}$, Shanghang Zhang$^{3}$, and Hao Dong$^{1,2}$
\thanks{*Equal contribution; $^{1}$CFCS, School of Computer Science, Peking University;
$^{2}$PKU-Agibot Lab
$^{3}$State Key Laboratory of Multimedia Information Processing, School of Computer Science, Peking University;
$^{4}$Tencent Robotics X Laboratory}
}

\begin{document}
\newcommand{\liang}[1]{\textcolor{red}{liang: #1}}

\maketitle
\thispagestyle{empty}
\pagestyle{empty}




\begin{abstract}
Recent advancements in imitation learning have shown promising results in robotic manipulation, driven by the availability of high-quality training data. 
To improve data collection efficiency, some approaches focus on developing specialized teleoperation devices for robot control, while others directly use human hand demonstrations to obtain training data. 
However, the former requires both a robotic system and a skilled operator, limiting scalability, while the latter faces challenges in aligning the visual gap between human hand demonstrations and the deployed robot observations. 
To address this, we propose a human hand data collection system combined with our hand-to-gripper generative model, which translates human hand demonstrations into robot gripper demonstrations, effectively bridging the observation gap.
Specifically, a GoPro fisheye camera is mounted on the human wrist to capture human hand demonstrations.
We then train a generative model on a self-collected dataset of paired human hand and UMI gripper demonstrations, which have been processed using a tailored data pre-processing strategy to ensure alignment in both timestamps and observations.
Therefore, given only human hand demonstrations, we are able to automatically extract the corresponding SE(3) actions and integrate them with high-quality generated robot demonstrations through our generation pipeline for training robotic policy model.
In experiments, the robust manipulation performance demonstrates not only the quality of the generated robot demonstrations but also the efficiency and practicality of our data collection method.
More demonstrations can be found at: \url{https://rwor.github.io/}.

\end{abstract}

\section{INTRODUCTION}

Imitation learning~\cite{chi2023diffusion,goyal2024rvt,ke20243d,ze20243d,gervet2023act3d,shridhar2023perceiver,shridhar2022cliport,torabi2018behavioral,zitkovich2023rt,jia2024lift3d,li20253dwg,yang2025lidar} with human demonstrations has advanced significantly in the field of robotic manipulation in recent years, leading to the development of various techniques for gathering such data. 
A widely used method is kinesthetic teaching, where a human manually guides the robot through a desired trajectory~\cite{argall2009survey}. 
While straightforward, this process is often slow and cumbersome. 
Another approach is teleoperation, where devices such as keyboards, video game controllers, or VR controllers are used to control the robot remotely~\cite{ding2024bunny,wu2024gello,arunachalam2023holo,handa2020dexpilot,qin2023anyteleop,song2020grasping,laghi2018shared,wu2019teleoperation}. 
However, these methods typically require both a robotic system and a skilled operator, which limits their scalability and ease of access for data collection.
\begin{figure}[htbp]
    \centering
    \includegraphics[width=0.50\textwidth]{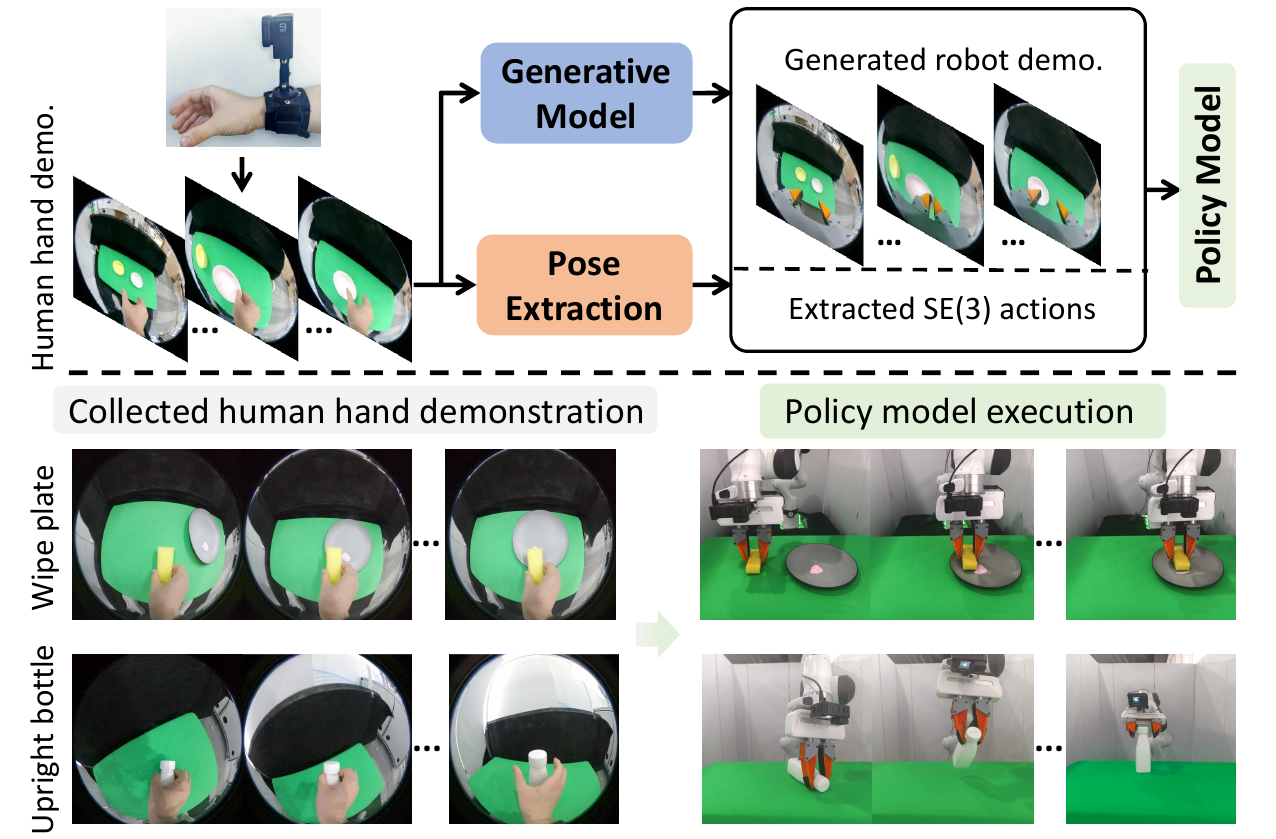}
    \vspace{-0.2cm}
    \caption{\textbf{RwoR's data collection pipeline for policy model training}. The top part illustrates the RwoR pipeline, which extracts actions from human hand demonstrations and trains a generative model to convert these demonstrations into UMI gripper demonstrations for policy learning. The bottom part visualizes the collected human hand demonstrations and the corresponding agent execution during real robot deployment.}
    \label{fig:teaser}
    \vspace{-0.6cm}
\end{figure}

Alternatively, human demonstrations can be collected using portable systems without the need for physical robot hardware~\cite{chen2024arcap,duan2023ar2,chi2024universal,wang2024dexcap,yang2022learning,xiong2023robotube,grauman2022ego4d,peng2024learning}.
On one hand, typical works such as UMI~\cite{chi2024universal}, use hand-held grippers paired with carefully designed interfaces to enable portable data collection, however, it still requires additional effort in terms of device setup.
On the other hand, other works~\cite{wang2024dexcap,duan2023ar2} focus on capturing motion actions from human hand demonstrations, which offers flexibility and efficiency. 
They then align the visual observation gap between human demonstrations (e.g., human hand) and real robot deployment (e.g., robot gripper) through techniques such as observation post-processing (e.g., editing and alignment) or AR rendering.
However, these manual adjustments or specific rule-based approaches face challenges in scalability due to their dependence on heuristic strategies, which are labor-intensive and may struggle to generalize across different environments or tasks.

Therefore, as shown in Fig.~\ref{fig:teaser}, our key insight is to develop a scalable and flexible data collection system based on human hand demonstrations that is not affected by the visual domain gap between human hand demonstrations and real robot observations.
Specifically, inspired by UMI~\cite{chi2024universal}, we attach a GoPro fisheye camera on the human wrist to record human hand demonstration videos. 
However, since the demonstration captures human hand rather than the robot gripper, directly training on this data would introduce a real robot test-time visual domain gap.
To tackle this challenge, we design a pipeline that collects paired human hand and UMI gripper demonstration and aligns them based on timestamps and observations, enabling the training of a hand-to-gripper generative model.
Therefore, leveraging this generative model, given human hand demonstrations, we first extract the hand pose to determine the corresponding parallel-jaw gripper SE(3) actions and then apply the hand-to-gripper generative model to convert the collected human hand demonstrations into robot gripper demonstrations, facilitating effective robotic policy model training.

Empirically, we leverage these generated demonstrations to train a policy model, which achieves similar performance compared with training on demonstrations collected using hand-held UMI grippers.
This validates the effectiveness and quality of the generated robot demonstrations for policy learning.
We then assess the quality of the generated robot demonstration video, which shows promising performance even on tasks and background scenes that are different from the limited self-collected generative model's training dataset.

In summary, our contributions are as follows:
\begin{itemize}
    \item We develop an efficient data collection system, RwoR, which leverages a trained hand-to-gripper generative model to transform collected human hand demonstrations into high-quality UMI gripper demonstrations, enabling effective policy learning.
    \item RwoR demonstrates comparable performance across a range of robotic manipulation tasks when compared to UMI, which trains on data collected using a hand-held gripper device.
    
\end{itemize}
\section{RELATED WORK}

\textbf{Learning from Demonstrations.}
Imitation Learning (IL) has emerged as a powerful paradigm for enabling robots to perform complex manipulation tasks by leveraging demonstrations provided by human experts.
Recent advancements in IL have leveraged deep neural networks to learn policies directly from raw image inputs~\cite{mandlekar2021matters, zhu2022viola,li2024manipllm,huang2024manipvqa,liu2024robomamba,li2025crayonrobo,liu2025hybridvla,chen2025fast,xu2024naturalvlm,cai2024spatialbot,xiong2024autonomous}, allowing robots to perform tasks even in complex environments with bimanual manipulators~\cite{chi2023diffusion,goyal2024rvt,torabi2018behavioral,zitkovich2023rt,zhao2023learning}. However, these methods largely depend on large-scale training data, making the collection of high-quality demonstration data particularly important.

\textbf{Demonstration collection methods.}
Demonstration collection is a crucial step in imitation learning, as the quality and quantity of demonstrations significantly impact the performance of the trained robot policies. 
Teleoperation techniques~\cite{ke2020telemanipulation, wang2021generalization, zhu2022viola, brohan2022rt, wu2023gello, gao2024efficient, lin2024learning, zhao2023learning, mobilealoha, opentv, cp, he2024omnih2o, ding2024bunny, qin2023anyteleop, van2024puppeteer} represent widely used categories of demonstration collection methods and can be broadly categorized into two types: teleoperation with robots and teleoperation without robots.
The first category involves leveraging various user interfaces such as keyboards, mice~\cite{kent2017comparison, leeper2012strategies}, video game controllers~\cite{laskey2017comparing} [15], 3D mice~\cite{dragan2012online, shridhar2023perceiver}, mobile phones~\cite{mandlekar2018roboturk} [28], and virtual reality (VR) controllers~\cite{whitney2019comparing, zhang2018deep, lipton2017baxter, jaegle2021perceiver,ding2024bunny}. These methods provide precise control over the robot and allow for the collection of high-quality, in-domain data. 
However, they require access to a physical robot, limiting scalability and accessibility. 
The second category focuses on collecting demonstrations without the need for physical robots. 
They either utilize a hand-held gripper~\cite{chi2024universal} or capture human hand movements, employing data retargeting~\cite{wang2024dexcap} or AR rendering techniques~\cite{duan2023ar2} to bridge the observation gap between human hand demonstrations and robot deployment.

Following this trend, our approach captures human hand demonstrations using a wrist-mounted camera and leverages a generative model to convert these into robot gripper demonstrations. This strategy effectively bridges the visual domain gap between human and robot observations, enabling scalable, low-cost data collection for imitation learning without the need for physical robot hardware.

\textbf{Image inpainting in robotics.}
Recent work has explored image inpainting and augmentation~\cite{chen2023genaug, mandi2022cacti, yu2023scaling,hirose2023exaug,bahl2022human} to generate a robot execution process. 
AR2-D2~\cite{duan2023ar2} uses augmented reality to replace human hands with virtual robot arms. 
Mirage~\cite{chen2024mirage} further improves this by using cross-painting to replace the target robot with the source robot in real time, creating the illusion that the source robot is performing the task. 
Unlike existing work, we enable a diffusion-based generative  model to transfer human hand to robot gripper demonstrations.
\begin{figure*}[htbp]
    \centering
    \includegraphics[width=\textwidth]{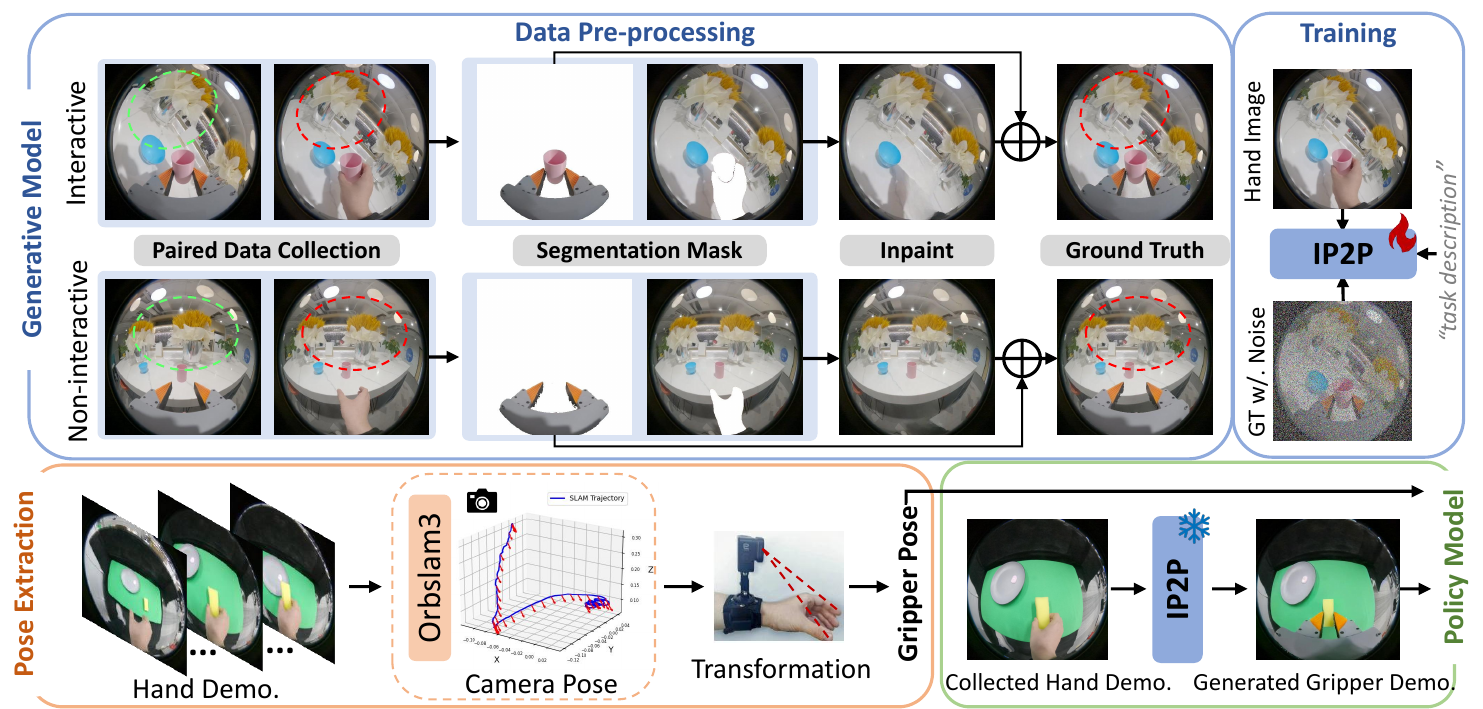}
    \vspace{-0.4cm}
    \caption{\textbf{Overall Pipeline}. We design a data pre-processing strategy to train the hand-to-gripper generative model. With this generative model, given human hand demonstrations, we first extract the corresponding gripper poses and then transform the demonstrations into gripper demonstrations using the generative model, thereby enabling effective training of the robotic policy model.}
    \label{fig:method}
    \vspace{-0.6cm}
\end{figure*}

\section{METHOD}

We begin by detailing the process of collecting human hand demonstrations in Sec. \ref{sec:method-setup}. In Sec. \ref{sec:method-diff}, we explain how we enable a generative model to convert human hand demonstrations into UMI gripper demonstrations, bridging the visual observation gap between human-collected data and real robot deployment. Finally, in Sec. \ref{sec:method-pl}, we apply an imitation learning policy that trains on the generated demonstrations.
\subsection{Human Demonstration Setup}
\label{sec:method-setup}
As shown in Fig.~\ref{fig:teaser}, inspired by UMI~\cite{chi2024universal}, we utilize a GoPro Hero9 camera paired with the GoPro Max Lens Mod 1.0, a fisheye lens that provides a wide field of view, to capture human hand demonstration.
The camera is mounted on a wrist with a corresponding adapter.
Since human hand demonstrations are later converted into UMI gripper demonstrations using our trained generative model for imitation policy learning, it is essential to ensure that the camera views used to capture the human hand demonstrations align with the wrist camera view on the deployed robot.
Therefore, the camera pose, including the rotation angle and distance from the fingertips, is carefully adjusted to maintain alignment between the two perspectives, aiming to minimize the potential visual gap between the human hand and the robot gripper.

\subsection{Generative model training}
\label{sec:method-diff}
Directly training a policy model using human hand demonstration videos is not feasible due to the observation gap between the human hand and the deployed robot. Existing methods often rely on augmented reality (AR) systems~\cite{duan2023ar2} or other rule-based techniques~\cite{wang2024dexcap} to replace the human hand with a robot for policy model training. However, these approaches lack scalability across different environments and tasks.
To address this challenge, our goal is to leverage the power of diffusion-based generative models~\cite{brooks2023instructpix2pix,rombach2022high} to transfer human hand demonstrations into robot gripper demonstrations, enabling more efficient and scalable visual gap transfer. However, due to the limited availability of paired human hand and robot gripper demonstrations, directly training the generative model is not feasible.

To resolve these, in Sec.~\ref{sec:diff-data}, we collect a set of paired human hand and robot gripper demonstration data and design pre-processing mechanism to align frames in terms of timestamp and observation.
Based on these paired data, in Sec.~\ref{sec:diff-train}, we train the generative model to effectively generate robot gripper demonstrations conditioned on the corresponding human hand demonstrations.

\subsubsection{Data Collection}
\label{sec:diff-data}

Since UMI~\cite{chi2024universal} offers a portable and flexible hand-held gripper for collecting robot demonstrations, it provides a practical solution to collect paired human hand and robot gripper demonstrations.

Specifically, as outlined in Sec. \ref{sec:method-setup}, we collect a human hand demonstration, denoted as $\mathcal{H} = \{h_1, \dots, h_{T_1}\}$, where $T_1$ is the total number of frames in that demonstration.
We then follow UMI' setup, which is mainly based on open-source 3D-printed design, to collect corresponding robot gripper demonstration $\mathcal{R} = \{r_1, \dots, r_{T_2}\}$,  $T_2$ is the number of robot gripper frames.
Since there is inevitable \textbf{timestamp misalignment}, we use Temporal Cycle-Consistency Learning (TCC)~\cite{Dwibedi_2019_CVPR}, which learns self-supervised representations to perform alignment between frames. We first train the model by inputting paired demonstration of human hand $\mathcal{H}$ and UMI gripper $\mathcal{R}$, obtaining the embeddings via cycle-consistency losses. The aligned human hand frames $\{h_1, \dots, h_{T}\}$ and UMI gripper frames $\{r_1, \dots, r_{T}\}$ of $T$ timestamp can subsequently be extracted through nearest neighbor retrieval in the embedding space. 
\begin{figure*}[htbp]
    \centering
    \includegraphics[width=\textwidth]{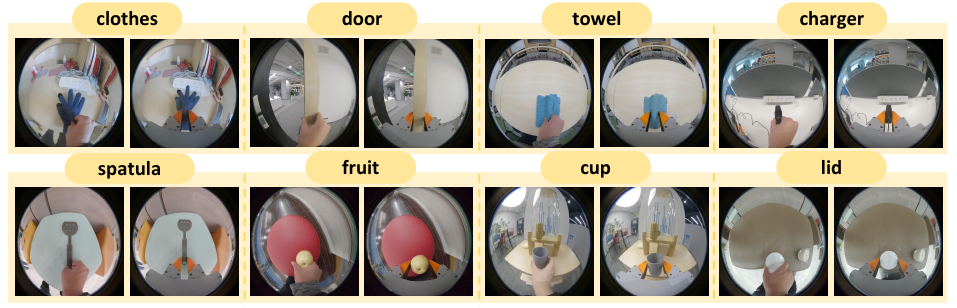}
    \caption{\textbf{Training Dataset for Generative model.} We visualize our self-collected dataset of paired human hand and UMI gripper demonstrations. The dataset includes a variety of household objects across different scenarios.}
    \label{fig:dataset}
    \vspace{-0.3cm}
\end{figure*}

Though we align the paired human hand and UMI gripper demonstrations based on timestamps, inevitable \textbf{observation misalignments} between human hand frame $h_t$ and robot gripper frame $r_t$ of same timestamp $t$ persist due to factors like inconsistent camera angles and discrepancies in execution trajectories when collecting demonstrations. 
In Fig. \ref{fig:method}, the red circle and green circle highlight the inconsistency observation gap between the human hand and umi gripper demonstrations.
As a result, training directly on these inconsistencies can distract the generative model, making it harder to learn precise and effective transformations.
However, our goal for the generative model is to transform the interaction between the human hand and the object into the corresponding interaction with the robot gripper, while keeping the background region unchanged, identical to that of the human hand demonstration.
To achieve this objective, we develop a data pre-processing strategy that refines the collected ground-truth robot gripper demonstrations, ensuring they differ from human hand demonstrations only in the foreground objects while preserving the background. This consistency facilitates effective learning for the generative model.

Specifically, as shown in Fig.~\ref{fig:method}, we automatically classify each frame into interactive stages and non-interactive stages based on the relative positions of the gripper and the interactive object.
During interactions, both the gripper and the interacting object are treated as foreground elements that require transformation.
In contrast, when no interaction occurs, only the gripper is considered a foreground object, while the interactive object is regarded as part of the background.
We then use SAM2~\cite{ravi2024sam} to extract the background $h_{t}^{b}$ of the human hand frame and the foreground objects $r_{t}^{f}$ from the corresponding robot gripper frame.
With the foreground and background segmentation mask, for both stages, we apply Inpaint Anything~\cite{yu2023inpaint} to inpaint the objects on the human hand background image $h_{t}^{b}$ and add the corresponding foreground objects from the robot gripper image $r_{t}^{f}$, forming a new ground-truth gripper demonstration $\hat{r_{t}}$ which is consistent with human hand demonstrations in terms of camera view and background.

In total, our self-collected dataset consists of 200 paired human hand and UMI gripper demonstrations, capturing interactions across a wide range of scenes, actions, and objects. The scenes include various lighting conditions and environments such as offices, classrooms, and home settings, while the actions cover tasks such as grasping, pushing, pulling, and rotating. The objects cover 25 categories, including household items, tools, and containers, with a total of 60 instances to ensure diverse variations in object appearance.
We visualize the collected dataset pairs in Fig.~\ref{fig:dataset}.
\subsubsection{Training Strategy}
\label{sec:diff-train}

 




For training the generative model $\epsilon_\theta$, we utilize a diffusion-based network, InstructPix2Pix (IP2P)\cite{brooks2023instructpix2pix}, which is built upon the Stable Diffusion model\cite{rombach2022high}.
We initialize the network with its pre-trained weights, enabling effective image-to-image transformation.
The model inputs consist of images from the human hand demonstrations $h_t$ and text input $l$ specifying the task description, such as ``Turn the hand into a gripper. The gripper is holding a \texttt{\{obj\_name\}}," where \texttt{\{obj\_name\}} represents the object being interacted.
The model’s output is supervised by the aligned robot gripper images $\hat{r_{t}}$.
We utilize the IP2P loss function, which is specifically designed for image-to-image transformation tasks in the context of prompt-based image generation.
The model learns to predict the noise added to the noisy latent space, conditioned on both human hand image and text input. 
Specifically, the loss function is defined as: 

\[
L = \mathbb{E}_{\mathcal{E}(\hat{r_{t}}), \mathcal{E}(h_t), l, \epsilon \sim \mathcal{N}(0,1), t} \left[ \| \epsilon - \epsilon_{\theta}(z_t, t, \mathcal{E}(h_t), l) \|^2_2 \right]
\]

, where \( z_t \) is the noisy latent representation at timestep \( t \), $\mathcal{E}$ is the encoder, $\epsilon_{\theta}$ is the diffusion generative network.
By doing so, the trained generative model learns to focus on the foreground region, ensuring a reliable transformation from human hand demonstrations to robot gripper demonstrations.



\subsection{Policy Model Training}
\label{sec:method-pl}
\subsubsection{Data Preparation}

During imitation learning policy training, the model typically requires \textbf{trajectory actions} and \textbf{corresponding robot desmontrations}. 
Specifically, for \textit{trajectory actions} extraction, after collecting human hand demonstration videos, we follow the pipeline used in UMI to extract end-effector position and rotation in SE(3) space.
We first utilize ORB-SLAM3~\cite{mur2015orb} for scene reconstruction and GoPro’s built-in IMU data for tracking to extract the camera's 6DoF pose. Then, using a fixed transformation matrix between the camera pose and the human hand's fingertip, we compute the fingertip’s 6DoF pose, which then serves as the robot gripper action for policy model training.
Compared to motion capture (MoCap) systems, this approach is more resilient to human hand occlusion, ensuring reliable and high-quality action extraction.
Regarding gripper open status, following the criteria in Sec.~\ref{sec:diff-data}, we use the relative positions of the object and human hand to determine if an interaction has occurred. 
If there is an interaction, the gripper is considered closed; otherwise, it is considered open.
For \textit{robot demonstrations} generation, we use the trained diffusion generative model to transfer the human hand demonstrations into UMI gripper demonstrations. 
This ensures each robot gripper frame is consistent with the human hand frame in terms of timestamp and visual observation, ensuring that the extracted actions can be aligned to robot demonstrations.

\begin{figure*}[htbp]
    \centering
    \includegraphics[width=\textwidth]{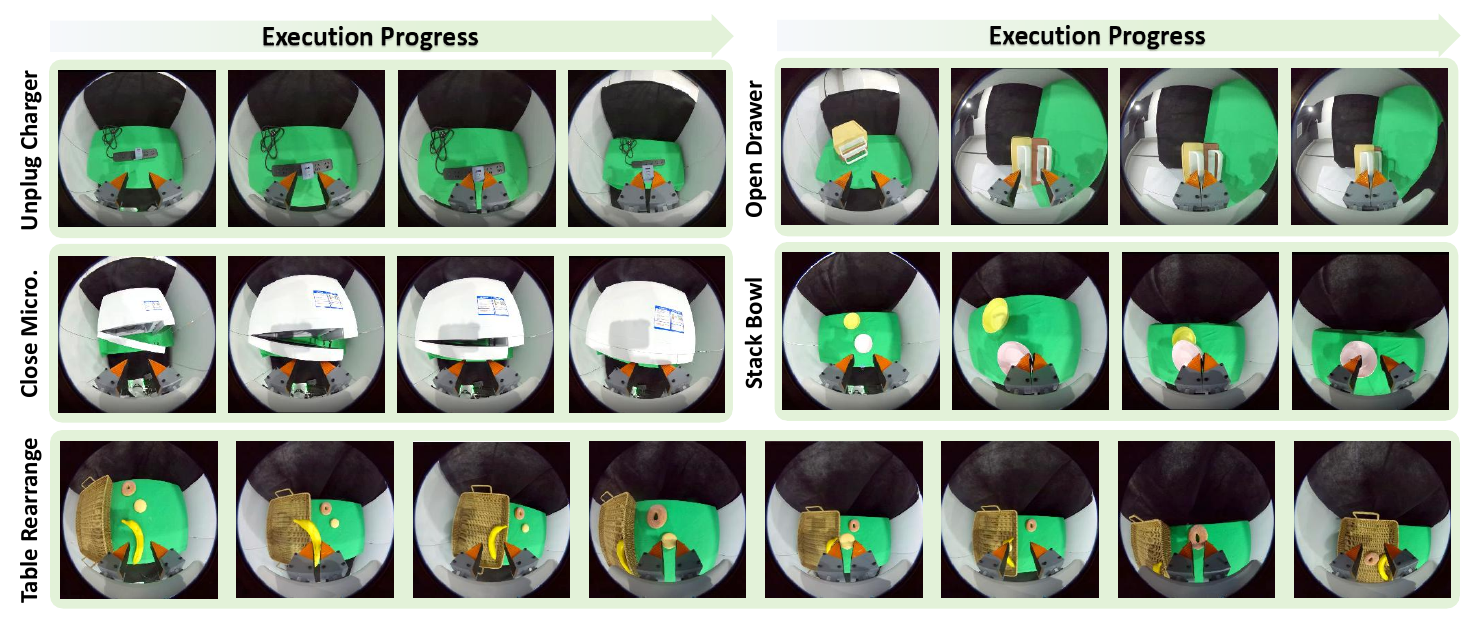}
    \caption{\textbf{Real-World Execution Visualization.} We visualize key frames of the agent's execution from the real-world manipulation perspective.}
    \label{fig:real-viz}
\end{figure*}
\begin{table*}[t]
\caption{\textbf{Real-World Performance.}  
We compare the success rate performance of the policy model when trained on datasets collected using our data collection framework versus those collected directly with UMI, with the latter serving as an upper bound. Additionally, we include a lower-bound baseline, `Alter.', for further comparison.
}
\centering
\small
\resizebox{\textwidth}{!}{
\begin{tabular}{lc|ccccccccc} 
\toprule
\rowcolor[HTML]{CBCEFB}

                                                & Avg.                 & Slide               & Pour     Water      &  Close    
                                                Micro. & Stack      Bowl  &    Upright Bottle   & Wipe Plate   & Unplug Charger& Open Drawer  
                                                & Table Rearrange
                                                \\

\midrule

UMI & \textbf{0.82}  & \textbf{0.93} & \textbf{0.87} & \textbf{1.0} & \textbf{0.80} & \textbf{0.80} & \textbf{0.73} & \textbf{0.83} & \textbf{0.77} & \textbf{0.66}\\ 
\rowcolor[HTML]{EFEFEF}Ours & 0.78 & \textbf{0.93} & \textbf{0.87} & 0.93 & 0.73 & 0.70  & 0.70 & 0.80 & \textbf{0.77} & 0.60  \\

Alter. & 0.37 & 0.43 & 0.40  & 0.53 & 0.33 & 0.37& 0.27 & 0.43 & 0.37 & 0.20 \\
\bottomrule
\end{tabular}
}

\label{tab:real}
\vspace{-0.4cm}
\end{table*}

\subsubsection{Policy Model Training}
Following UMI's~\cite{chi2024universal} policy interface design, we utilize the generated robot gripper demonstration data to train a visuomotor policy. 
This policy takes a sequence of generated UMI gripper images, the 6DoF end-effector pose, and the gripper open status as input, and outputs corresponding actions, including the end-effector pose and gripper open status.
In this work, we employ Diffusion Policy~\cite{chi2023diffusion} for all experiments.
By leveraging our RwoR, we are able to collect training data only with human hand demonstrations, without the need for precise camera calibration and a real robot system. 



\section{EXPERIMENT}
We evaluate the performance of the whole framework on a real robot with an imitation learning policy in Sec. \ref{sec:real-robot} and verify the quality of our generated videos in Sec. \ref{sec:exp-video}.

\subsection{Evaluation with Real Robot Deployment}
\label{sec:real-robot}
\subsubsection{Implementation Details}
During real robot deployment, our method is evaluated across 9 tasks on the Franka Research 3 (FR3) robot with a 3D-printed UMI gripper~\cite{chi2024universal}.  
We use a Gopro 9 camera to obtain real-world visual observations from the wrist view.  
For each task, 50 training human hand demonstrations are collected in a specific working space range.  
We train an agent for each task and evaluate each task in 15 trials within the training working space.
The success rate is used as the evaluation metric.

\subsubsection{Baseline Comparisons}
We adopt UMI~\cite{chi2024universal} as our comparison baseline, which utilizes a hand-held gripper to collect robot gripper demonstrations, serving as the upper bound.
The training data for UMI is collected within the same workspace range as ours, with the same number of samples and training epochs for diffusion policy training. 
As shown in Tab.~\ref{tab:real}, our framework achieves a similar success rate compared to UMI, demonstrating that the quality of our generated robot gripper demonstrations is comparable to those collected directly using a hand-held device.
We visualize the keyframes of the execution process in Fig.~\ref{fig:real-viz}, where the \textit{table rearrangement} task requires the agent to sequentially pick up three objects and place them into the container.
Note that since our diffusion generative model primarily learns to transfer foreground regions and remains robust to background variations, it can effectively handle manipulation scenes that differ from those in the generative model’s training dataset (Fig.~\ref{fig:dataset}).

Moreover, we further compare our method with a baseline (the third row in Tab.\ref{tab:real}), which uses an alternative approach for generating training ground truth.
Specifically, we replace our approach of using a generative model to transfer human hand demonstrations to UMI gripper demonstrations with a simple rule-based texture mapping method. 
In this approach, we remove the hand parts from the human hand demonstration, inpaint the background, and then apply the UMI gripper pattern to the image. 
These modified UMI gripper demonstrations are then used to train the policy model while keeping all other training parameters unchanged. 
As a result, the model trained using this data experiences a significant performance drop. 
This occurs because the rule-based texture mapping introduces inaccuracies in the relative positioning between the gripper and the object, creating a significant visual observation gap. 
This gap further impacts the policy model’s learning, resulting in issues such as stagnation, failure to advance toward the target object, and improper timing of gripper opening.


\begin{figure*}[t]
    \centering
    \includegraphics[width=\textwidth]{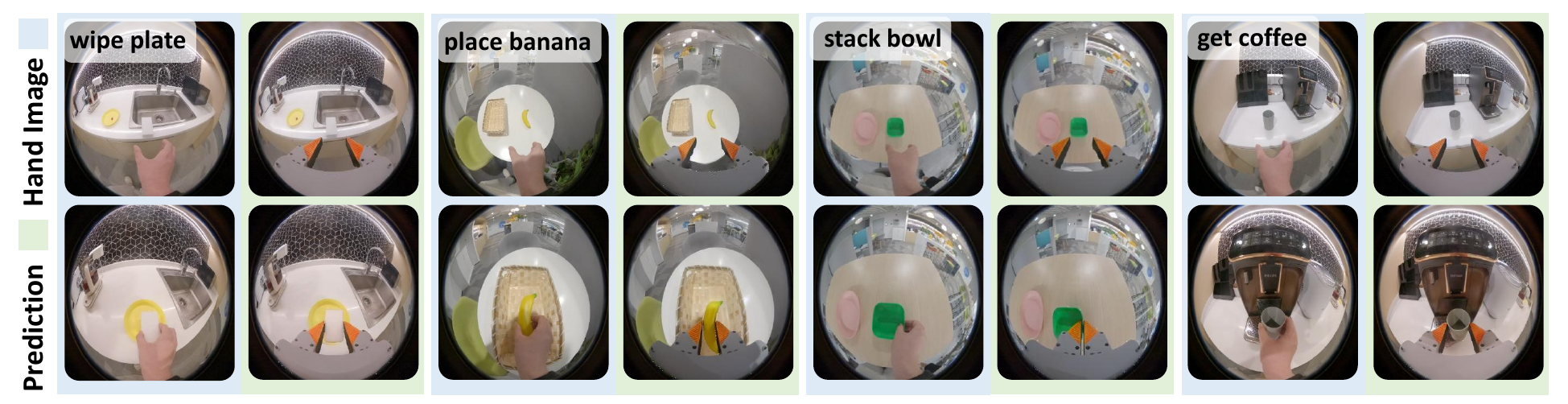}
    \caption{\textbf{Visualization of generative model's performance.} We visualize the input human hand image provided to the generative model alongside its corresponding predicted UMI gripper image.
}
    \label{fig:real-gen}
    \vspace{-0.3cm}
\end{figure*}
\begin{table}
 \centering
 \caption{Generative model's performance and ablation study.}
\normalsize
 \begin{tabular}{lcccc}
\toprule
\rowcolor[HTML]{CBCEFB}
Row ID                                     & Edit     & Interactive   & PSNR$\uparrow$                    & SSIM$\uparrow$ \\
\midrule
1       & \no & \no & 31.53 & 0.77   \\
\rowcolor[HTML]{EFEFEF}
2    &\yes & \no & 33.37 & 0.84    \\
3(Ours)   &\yes&\yes & \textbf{33.80} & \textbf{0.86}\\
\bottomrule
 \end{tabular}

 \label{tab:gen-abla}
 \vspace{-0.2cm}
\end{table}
\subsection{Evaluation of Images Transferring}
\label{sec:exp-video}
\subsubsection{Setting Details}
We train our diffusion generative model on 200 paired human hand and UMI gripper demonstrations with 15,000 extracted frames, encompassing a diverse range of scenes, actions, and interactive objects. 
For testing, we evaluate the model on 2000 frames from the training demonstrations but are distinct from those frames used during training, ensuring that the model is assessed on unseen samples. 
To measure the quality of the generated robot gripper images, we use PSNR and SSIM as evaluation metrics to calculate the similarity between the predicted images and ground-truth gripper images $\hat{r_t}$. 
\subsubsection{Result Analysis}

In Tab.~\ref{tab:gen-abla} Row 3, the trained model achieves promising performance measured by PSNR and SSIM.
The model is trained on ground truth generated through our designed data collection and pre-processing mechanism. We visualize the generated robot gripper image and corresponding human hand image input in Fig.~\ref{fig:real-gen}.

To verify the effectiveness of the proposed components in ground truth generation, we conduct the following ablation study:
In Tab.~\ref{tab:gen-abla} Row 1, we use the original timestamp-aligned robot gripper images as the ground truth, ignoring the observation misalignment processing. 
This leads to a performance drop of 2.33 in PSNR and 0.09 in SSIM, highlighting the importance of addressing observation misalignment caused by subtle discrepancies in camera view angles and trajectory movements.
In Tab.~\ref{tab:gen-abla} Row 2, 
we modify the observation from the robot gripper demonstration but do not differentiate between the interactive and non-interactive stages as ours. 
Specifically, for all frames, we remain the gripper from the robot gripper demonstration, while interactive objects and background regions come from human hand demonstrations.
Compared to our method, this also results in performance degradation.
Although the margin is minimal, this is because these metrics do not fully capture the plausibility of the generated video.
However, through manual inspection, we observe that during the interaction, there are numerous inconsistencies in the relationship between the interactive object and gripper, which can confuse the policy model’s learning.


\subsubsection{Generalization study}
\begin{figure}[h]
    \centering
    \includegraphics[width=\linewidth]{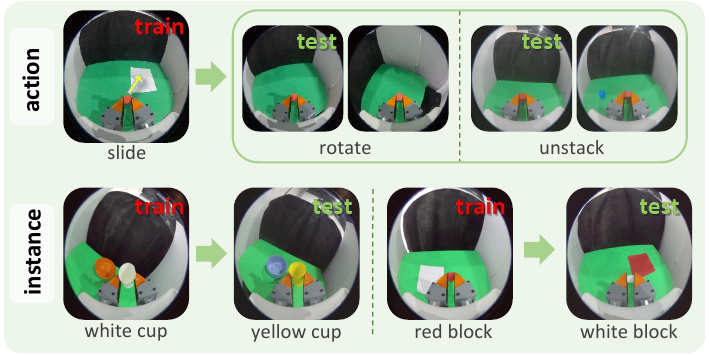}  

    \caption{Visualization of generative model's generalization ability on unseen action types and instances.}
    \vspace{-0.2cm}
    \label{fig:generative}
\end{figure}

As shown in Fig.\ref{fig:generative}, we use the generated robot demonstrations to train the policy model and test its success rate, which can reflect the quality of generated robot gripper demonstrations on unseen \textit{action types} and \textit{instances}.
For \textbf{action type generalization}, as shown in Tab.\ref{tab:generative} Ex1, the generative model’s training dataset includes actions ``slide block", however, we use the trained diffusion-based generative model to generate robot demonstrations for other actions, such as ``rotate block" and ``unstack block."
Even for unseen action tasks, the success rate of the trained policy model remains high, demonstrating that the generative model produces high-quality robot demonstrations for novel actions.
This is because our generative model focuses on transforming the foreground objects in human hand demonstrations into robot grippers, rather than depending on the specific action being performed, thus allowing it to generalize across various action types.
For \textbf{instance generalization}, we test the trained generative model on unseen instance appearance, which is not included in the entire generative model training dataset.
Comparing Tab.\ref{tab:generative} Ex2 and Tab.\ref{tab:real}, the success rate remains consistent at 0.87 for both the seen instance pouring the ``white cup'' and the unseen instance pouring the ``yellow cup''. Additionally, the unseen instance sliding the ``white block'' experiences only a minor performance drop of 0.04 compared to the seen instance sliding the ``red block''.
These results demonstrate that our generative model can effectively transform previously unseen object instances, generating high-quality demonstrations for policy model training.






\begin{table}[t]
 \centering
 \caption{Success rate of the policy model on action types (Ex1.) and instances (Ex2.) that are unseen to generative model.}

\small
 \begin{tabular}{ccc|ccc}
\toprule
\rowcolor[HTML]{CBCEFB}
Ex1.                        &          Rotate & Unstack  & Ex2.& \textcolor{yellow}{Pour water}&\textcolor{white}{Slide block}\\
\midrule
&0.80 &0.83 &  & 0.87 & 0.87  \\
\bottomrule
 \end{tabular}

\vspace{-0.2cm}
 
 \label{tab:generative}
\end{table}


\section{CONCLUSIONS}
We develop a robust and efficient data collection system that leverages human hand demonstrations to generate high-quality UMI gripper demonstrations, enabling effective policy training. 
By collecting 200 paired human hand and robot demonstrations, we train a diffusion-based generative model that successfully bridges the observation gap between human hand data and real robot observations. 
Our RwoR framework demonstrates comparable performance across various robotic manipulation tasks when compared to UMI, which is trained on data collected using a hand-held gripper device. 




\section{Limitations and Future Work}
\label{sec:app-fail}
The typical failure modes of the framework can be categorized into two main types: a) Poor generated video quality, and b) Exceeding joint limitations.

\textit{a) Poor generated video quality.}
Since the diffusion policy (DP) is a purely visual-motor policy model, the quality of the training data—specifically the generated robot gripper demonstrations—is crucial for the success of the model. 
For instance, if the generated video shows the gripper closing before making contact with the object, during deployment to a real robot, the trained policy model may close before contacting. 
As a result, poor quality in the generated demonstrations directly affects the learning process of the policy model, potentially leading to suboptimal or incorrect behavior during real-world manipulation. 

\textit{b) Exceeding joint limitations.}
Since the data is collected from human hand demonstrations, and the degrees of freedom of the human hand and arm are greater than those of a robot arm, some trajectories captured by the human hand may be outside the robot’s reach or capabilities. 
Therefore, it is essential to ensure that the actions being demonstrated are suitable for the robot arm’s limitations during data collection. 

\textit{Limitations.}
Since the training data for the generative model is self-collected, the range of interacting objects is limited, and the data may lack strong instance-level generalization. 
As a result, when training the diffusion policy, it is crucial to interact with objects that are similar to those seen in the training set for the generative model.
However, for similar objects, variations in actions, camera views, and backgrounds can still be accommodated. 
Therefore, we believe that as the diversity of paired training data increases, the generative model's generalization ability will gradually improve, enabling it to handle a broader range of robotic tasks.


\section*{Acknowledgement}
This project was supported by the National Youth Talent Support Program (8200800081), the National Natural Science Foundation of China (No. 62376006 and No. 62136001).


{
\bibliographystyle{IEEEtran}
\bibliography{IEEEabrv,reference}

\begin{thebibliography}{10}
\providecommand{\url}[1]{#1}
\csname url@rmstyle\endcsname
\providecommand{\newblock}{\relax}
\providecommand{\bibinfo}[2]{#2}
\providecommand\BIBentrySTDinterwordspacing{\spaceskip=0pt\relax}
\providecommand\BIBentryALTinterwordstretchfactor{4}
\providecommand\BIBentryALTinterwordspacing{\spaceskip=\fontdimen2\font plus
\BIBentryALTinterwordstretchfactor\fontdimen3\font minus \fontdimen4\font\relax}
\providecommand\BIBforeignlanguage[2]{{%
\expandafter\ifx\csname l@#1\endcsname\relax
\typeout{** WARNING: IEEEtran.bst: No hyphenation pattern has been}%
\typeout{** loaded for the language `#1'. Using the pattern for}%
\typeout{** the default language instead.}%
\else
\language=\csname l@#1\endcsname
\fi
#2}}

\bibitem{chi2023diffusion}
C.~Chi, Z.~Xu, S.~Feng, E.~Cousineau, Y.~Du, B.~Burchfiel, R.~Tedrake, and S.~Song, ``Diffusion policy: Visuomotor policy learning via action diffusion,'' \emph{The International Journal of Robotics Research}, p. 02783649241273668, 2023.

\bibitem{goyal2024rvt}
A.~Goyal, V.~Blukis, J.~Xu, Y.~Guo, Y.-W. Chao, and D.~Fox, ``Rvt-2: Learning precise manipulation from few demonstrations,'' \emph{arXiv preprint arXiv:2406.08545}, 2024.

\bibitem{ke20243d}
T.-W. Ke, N.~Gkanatsios, and K.~Fragkiadaki, ``3d diffuser actor: Policy diffusion with 3d scene representations,'' \emph{arXiv preprint arXiv:2402.10885}, 2024.

\bibitem{ze20243d}
Y.~Ze, G.~Zhang, K.~Zhang, C.~Hu, M.~Wang, and H.~Xu, ``3d diffusion policy: Generalizable visuomotor policy learning via simple 3d representations,'' in \emph{ICRA 2024 Workshop on 3D Visual Representations for Robot Manipulation}, 2024.

\bibitem{gervet2023act3d}
T.~Gervet, Z.~Xian, N.~Gkanatsios, and K.~Fragkiadaki, ``Act3d: 3d feature field transformers for multi-task robotic manipulation,'' in \emph{7th Annual Conference on Robot Learning}, 2023.

\bibitem{shridhar2023perceiver}
M.~Shridhar, L.~Manuelli, and D.~Fox, ``Perceiver-actor: A multi-task transformer for robotic manipulation,'' in \emph{Conference on Robot Learning}.\hskip 1em plus 0.5em minus 0.4em\relax PMLR, 2023, pp. 785--799.

\bibitem{shridhar2022cliport}
------, ``Cliport: What and where pathways for robotic manipulation,'' in \emph{Conference on robot learning}.\hskip 1em plus 0.5em minus 0.4em\relax PMLR, 2022, pp. 894--906.

\bibitem{torabi2018behavioral}
F.~Torabi, G.~Warnell, and P.~Stone, ``Behavioral cloning from observation,'' \emph{arXiv preprint arXiv:1805.01954}, 2018.

\bibitem{zitkovich2023rt}
B.~Zitkovich, T.~Yu, S.~Xu, P.~Xu, T.~Xiao, F.~Xia, J.~Wu, P.~Wohlhart, S.~Welker, A.~Wahid, \emph{et~al.}, ``Rt-2: Vision-language-action models transfer web knowledge to robotic control,'' in \emph{7th Annual Conference on Robot Learning}, 2023.

\bibitem{jia2024lift3d}
Y.~Jia, J.~Liu, S.~Chen, C.~Gu, Z.~Wang, L.~Luo, L.~Lee, P.~Wang, Z.~Wang, R.~Zhang, \emph{et~al.}, ``Lift3d foundation policy: Lifting 2d large-scale pretrained models for robust 3d robotic manipulation,'' \emph{arXiv preprint arXiv:2411.18623}, 2024.

\bibitem{li20253dwg}
X.~Li, J.~Liu, N.~Han, L.~Heng, Y.~Guo, H.~Dong, and Y.~Liu, ``3dwg: 3d weakly supervised visual grounding via category and instance-level alignment,'' \emph{arXiv preprint arXiv:2505.01809}, 2025.

\bibitem{yang2025lidar}
S.~Yang, J.~Liu, R.~Zhang, M.~Pan, Z.~Guo, X.~Li, Z.~Chen, P.~Gao, H.~Li, Y.~Guo, \emph{et~al.}, ``Lidar-llm: Exploring the potential of large language models for 3d lidar understanding,'' in \emph{Proceedings of the AAAI Conference on Artificial Intelligence}, vol.~39, no.~9, 2025, pp. 9247--9255.

\bibitem{argall2009survey}
B.~D. Argall, S.~Chernova, M.~Veloso, and B.~Browning, ``A survey of robot learning from demonstration,'' \emph{Robotics and autonomous systems}, vol.~57, no.~5, pp. 469--483, 2009.

\bibitem{ding2024bunny}
R.~Ding, Y.~Qin, J.~Zhu, C.~Jia, S.~Yang, R.~Yang, X.~Qi, and X.~Wang, ``Bunny-visionpro: Real-time bimanual dexterous teleoperation for imitation learning,'' \emph{arXiv preprint arXiv:2407.03162}, 2024.

\bibitem{wu2024gello}
P.~Wu, Y.~Shentu, Z.~Yi, X.~Lin, and P.~Abbeel, ``Gello: A general, low-cost, and intuitive teleoperation framework for robot manipulators,'' in \emph{2024 IEEE/RSJ International Conference on Intelligent Robots and Systems (IROS)}.\hskip 1em plus 0.5em minus 0.4em\relax IEEE, 2024, pp. 12\,156--12\,163.

\bibitem{arunachalam2023holo}
S.~P. Arunachalam, I.~G{\"u}zey, S.~Chintala, and L.~Pinto, ``Holo-dex: Teaching dexterity with immersive mixed reality,'' in \emph{2023 IEEE International Conference on Robotics and Automation (ICRA)}.\hskip 1em plus 0.5em minus 0.4em\relax IEEE, 2023, pp. 5962--5969.

\bibitem{handa2020dexpilot}
A.~Handa, K.~Van~Wyk, W.~Yang, J.~Liang, Y.-W. Chao, Q.~Wan, S.~Birchfield, N.~Ratliff, and D.~Fox, ``Dexpilot: Vision-based teleoperation of dexterous robotic hand-arm system,'' in \emph{2020 IEEE International Conference on Robotics and Automation (ICRA)}.\hskip 1em plus 0.5em minus 0.4em\relax IEEE, 2020, pp. 9164--9170.

\bibitem{qin2023anyteleop}
Y.~Qin, W.~Yang, B.~Huang, K.~Van~Wyk, H.~Su, X.~Wang, Y.-W. Chao, and D.~Fox, ``Anyteleop: A general vision-based dexterous robot arm-hand teleoperation system,'' \emph{arXiv preprint arXiv:2307.04577}, 2023.

\bibitem{song2020grasping}
S.~Song, A.~Zeng, J.~Lee, and T.~Funkhouser, ``Grasping in the wild: Learning 6dof closed-loop grasping from low-cost demonstrations,'' \emph{IEEE Robotics and Automation Letters}, vol.~5, no.~3, pp. 4978--4985, 2020.

\bibitem{laghi2018shared}
M.~Laghi, M.~Maimeri, M.~Marchand, C.~Leparoux, M.~Catalano, A.~Ajoudani, and A.~Bicchi, ``Shared-autonomy control for intuitive bimanual tele-manipulation,'' in \emph{2018 IEEE-RAS 18th International Conference on Humanoid Robots (Humanoids)}.\hskip 1em plus 0.5em minus 0.4em\relax IEEE, 2018, pp. 1--9.

\bibitem{wu2019teleoperation}
Y.~Wu, P.~Balatti, M.~Lorenzini, F.~Zhao, W.~Kim, and A.~Ajoudani, ``A teleoperation interface for loco-manipulation control of mobile collaborative robotic assistant,'' \emph{IEEE Robotics and Automation Letters}, vol.~4, no.~4, pp. 3593--3600, 2019.

\bibitem{chen2024arcap}
S.~Chen, C.~Wang, K.~Nguyen, L.~Fei-Fei, and C.~K. Liu, ``Arcap: Collecting high-quality human demonstrations for robot learning with augmented reality feedback,'' \emph{arXiv preprint arXiv:2410.08464}, 2024.

\bibitem{duan2023ar2}
J.~Duan, Y.~R. Wang, M.~Shridhar, D.~Fox, and R.~Krishna, ``Ar2-d2: Training a robot without a robot,'' \emph{arXiv preprint arXiv:2306.13818}, 2023.

\bibitem{chi2024universal}
C.~Chi, Z.~Xu, C.~Pan, E.~Cousineau, B.~Burchfiel, S.~Feng, R.~Tedrake, and S.~Song, ``Universal manipulation interface: In-the-wild robot teaching without in-the-wild robots,'' \emph{arXiv preprint arXiv:2402.10329}, 2024.

\bibitem{wang2024dexcap}
C.~Wang, H.~Shi, W.~Wang, R.~Zhang, L.~Fei-Fei, and C.~K. Liu, ``Dexcap: Scalable and portable mocap data collection system for dexterous manipulation,'' \emph{arXiv preprint arXiv:2403.07788}, 2024.

\bibitem{yang2022learning}
J.~Yang, J.~Zhang, C.~Settle, A.~Rai, R.~Antonova, and J.~Bohg, ``Learning periodic tasks from human demonstrations,'' in \emph{2022 International Conference on Robotics and Automation (ICRA)}.\hskip 1em plus 0.5em minus 0.4em\relax IEEE, 2022, pp. 8658--8665.

\bibitem{xiong2023robotube}
H.~Xiong, H.~Fu, J.~Zhang, C.~Bao, Q.~Zhang, Y.~Huang, W.~Xu, A.~Garg, and C.~Lu, ``Robotube: Learning household manipulation from human videos with simulated twin environments,'' in \emph{Conference on Robot Learning}.\hskip 1em plus 0.5em minus 0.4em\relax PMLR, 2023, pp. 1--10.

\bibitem{grauman2022ego4d}
K.~Grauman, A.~Westbury, E.~Byrne, Z.~Chavis, A.~Furnari, R.~Girdhar, J.~Hamburger, H.~Jiang, M.~Liu, X.~Liu, \emph{et~al.}, ``Ego4d: Around the world in 3,000 hours of egocentric video,'' in \emph{Proceedings of the IEEE/CVF conference on computer vision and pattern recognition}, 2022, pp. 18\,995--19\,012.

\bibitem{peng2024learning}
Z.~M. Peng, W.~Mo, C.~Duan, Q.~Li, and B.~Zhou, ``Learning from active human involvement through proxy value propagation,'' \emph{Advances in neural information processing systems}, vol.~36, 2024.

\bibitem{mandlekar2021matters}
A.~Mandlekar, D.~Xu, J.~Wong, S.~Nasiriany, C.~Wang, R.~Kulkarni, L.~Fei-Fei, S.~Savarese, Y.~Zhu, and R.~Mart{\'\i}n-Mart{\'\i}n, ``What matters in learning from offline human demonstrations for robot manipulation,'' \emph{arXiv preprint arXiv:2108.03298}, 2021.

\bibitem{zhu2022viola}
Y.~Zhu, A.~Joshi, P.~Stone, and Y.~Zhu, ``Viola: Imitation learning for vision-based manipulation with object proposal priors,'' \emph{arXiv preprint arXiv:2210.11339}, 2022.

\bibitem{li2024manipllm}
X.~Li, M.~Zhang, Y.~Geng, H.~Geng, Y.~Long, Y.~Shen, R.~Zhang, J.~Liu, and H.~Dong, ``Manipllm: Embodied multimodal large language model for object-centric robotic manipulation,'' in \emph{Proceedings of the IEEE/CVF Conference on Computer Vision and Pattern Recognition}, 2024, pp. 18\,061--18\,070.

\bibitem{huang2024manipvqa}
S.~Huang, I.~Ponomarenko, Z.~Jiang, X.~Li, X.~Hu, P.~Gao, H.~Li, and H.~Dong, ``Manipvqa: Injecting robotic affordance and physically grounded information into multi-modal large language models,'' in \emph{2024 IEEE/RSJ International Conference on Intelligent Robots and Systems (IROS)}.\hskip 1em plus 0.5em minus 0.4em\relax IEEE, 2024, pp. 7580--7587.

\bibitem{liu2024robomamba}
J.~Liu, M.~Liu, Z.~Wang, L.~Lee, K.~Zhou, P.~An, S.~Yang, R.~Zhang, Y.~Guo, and S.~Zhang, ``Robomamba: Multimodal state space model for efficient robot reasoning and manipulation,'' \emph{arXiv preprint arXiv:2406.04339}, 2024.

\bibitem{li2025crayonrobo}
X.~Li, L.~Xu, M.~Zhang, J.~Liu, Y.~Shen, I.~Ponomarenko, J.~Xu, L.~Heng, S.~Huang, S.~Zhang, \emph{et~al.}, ``Crayonrobo: Object-centric prompt-driven vision-language-action model for robotic manipulation,'' \emph{arXiv preprint arXiv:2505.02166}, 2025.

\bibitem{liu2025hybridvla}
J.~Liu, H.~Chen, P.~An, Z.~Liu, R.~Zhang, C.~Gu, X.~Li, Z.~Guo, S.~Chen, M.~Liu, \emph{et~al.}, ``Hybridvla: Collaborative diffusion and autoregression in a unified vision-language-action model,'' \emph{arXiv preprint arXiv:2503.10631}, 2025.

\bibitem{chen2025fast}
H.~Chen, J.~Liu, C.~Gu, Z.~Liu, R.~Zhang, X.~Li, X.~He, Y.~Guo, C.-W. Fu, S.~Zhang, \emph{et~al.}, ``Fast-in-slow: A dual-system foundation model unifying fast manipulation within slow reasoning,'' \emph{arXiv preprint arXiv:2506.01953}, 2025.

\bibitem{xu2024naturalvlm}
R.~Xu, Y.~Shen, X.~Li, R.~Wu, and H.~Dong, ``Naturalvlm: Leveraging fine-grained natural language for affordance-guided visual manipulation,'' \emph{IEEE Robotics and Automation Letters}, 2024.

\bibitem{cai2024spatialbot}
W.~Cai, I.~Ponomarenko, J.~Yuan, X.~Li, W.~Yang, H.~Dong, and B.~Zhao, ``Spatialbot: Precise spatial understanding with vision language models,'' \emph{arXiv preprint arXiv:2406.13642}, 2024.

\bibitem{xiong2024autonomous}
C.~Xiong, C.~Shen, X.~Li, K.~Zhou, J.~Liu, R.~Wang, and H.~Dong, ``Autonomous interactive correction mllm for robust robotic manipulation,'' in \emph{8th Annual Conference on Robot Learning}, 2024.

\bibitem{zhao2023learning}
T.~Z. Zhao, V.~Kumar, S.~Levine, and C.~Finn, ``Learning fine-grained bimanual manipulation with low-cost hardware,'' \emph{arXiv preprint arXiv:2304.13705}, 2023.

\bibitem{ke2020telemanipulation}
L.~Ke, A.~Kamat, J.~Wang, T.~Bhattacharjee, C.~Mavrogiannis, and S.~S. Srinivasa, ``Telemanipulation with chopsticks: Analyzing human factors in user demonstrations,'' in \emph{2020 IEEE/RSJ International Conference on Intelligent Robots and Systems (IROS)}.\hskip 1em plus 0.5em minus 0.4em\relax IEEE, 2020, pp. 11\,539--11\,546.

\bibitem{wang2021generalization}
C.~Wang, R.~Wang, A.~Mandlekar, L.~Fei-Fei, S.~Savarese, and D.~Xu, ``Generalization through hand-eye coordination: An action space for learning spatially-invariant visuomotor control,'' in \emph{2021 IEEE/RSJ International Conference on Intelligent Robots and Systems (IROS)}.\hskip 1em plus 0.5em minus 0.4em\relax IEEE, 2021, pp. 8913--8920.

\bibitem{brohan2022rt}
A.~Brohan, N.~Brown, J.~Carbajal, Y.~Chebotar, J.~Dabis, C.~Finn, K.~Gopalakrishnan, K.~Hausman, A.~Herzog, J.~Hsu, \emph{et~al.}, ``Rt-1: Robotics transformer for real-world control at scale,'' \emph{arXiv preprint arXiv:2212.06817}, 2022.

\bibitem{wu2023gello}
P.~Wu, Y.~Shentu, Z.~Yi, X.~Lin, and P.~Abbeel, ``Gello: A general, low-cost, and intuitive teleoperation framework for robot manipulators,'' \emph{arXiv preprint arXiv:2309.13037}, 2023.

\bibitem{gao2024efficient}
J.~Gao, A.~Xie, T.~Xiao, C.~Finn, and D.~Sadigh, ``Efficient data collection for robotic manipulation via compositional generalization,'' \emph{arXiv preprint arXiv:2403.05110}, 2024.

\bibitem{lin2024learning}
T.~Lin, Y.~Zhang, Q.~Li, H.~Qi, B.~Yi, S.~Levine, and J.~Malik, ``Learning visuotactile skills with two multifingered hands,'' \emph{arXiv:2404.16823}, 2024.

\bibitem{mobilealoha}
Z.~Fu, T.~Z. Zhao, and C.~Finn, ``Mobile aloha: Learning bimanual mobile manipulation with low-cost whole-body teleoperation,'' \emph{arXiv preprint arXiv:2401.02117}, 2024.

\bibitem{opentv}
X.~Cheng, J.~Li, S.~Yang, G.~Yang, and X.~Wang, ``Open-television: teleoperation with immersive active visual feedback,'' \emph{arXiv preprint arXiv:2407.01512}, 2024.

\bibitem{cp}
A.~Prasad, K.~Lin, J.~Wu, L.~Zhou, and J.~Bohg, ``Consistency policy: Accelerated visuomotor policies via consistency distillation,'' \emph{arXiv preprint arXiv:2405.07503}, 2024.

\bibitem{he2024omnih2o}
T.~He, Z.~Luo, X.~He, W.~Xiao, C.~Zhang, W.~Zhang, K.~Kitani, C.~Liu, and G.~Shi, ``Omnih2o: Universal and dexterous human-to-humanoid whole-body teleoperation and learning,'' \emph{arXiv preprint arXiv:2406.08858}, 2024.

\bibitem{van2024puppeteer}
J.~van Haastregt, M.~C. Welle, Y.~Zhang, and D.~Kragic, ``Puppeteer your robot: Augmented reality leader-follower teleoperation,'' \emph{arXiv preprint arXiv:2407.11741}, 2024.

\bibitem{kent2017comparison}
D.~Kent, C.~Saldanha, and S.~Chernova, ``A comparison of remote robot teleoperation interfaces for general object manipulation,'' in \emph{Proceedings of the 2017 ACM/IEEE international conference on human-robot interaction}, 2017, pp. 371--379.

\bibitem{leeper2012strategies}
A.~E. Leeper, K.~Hsiao, M.~Ciocarlie, L.~Takayama, and D.~Gossow, ``Strategies for human-in-the-loop robotic grasping,'' in \emph{Proceedings of the seventh annual ACM/IEEE international conference on Human-Robot Interaction}, 2012, pp. 1--8.

\bibitem{laskey2017comparing}
M.~Laskey, C.~Chuck, J.~Lee, J.~Mahler, S.~Krishnan, K.~Jamieson, A.~Dragan, and K.~Goldberg, ``Comparing human-centric and robot-centric sampling for robot deep learning from demonstrations,'' in \emph{2017 IEEE International Conference on Robotics and Automation (ICRA)}.\hskip 1em plus 0.5em minus 0.4em\relax IEEE, 2017, pp. 358--365.

\bibitem{dragan2012online}
A.~D. Dragan and S.~S. Srinivasa, ``Online customization of teleoperation interfaces,'' in \emph{2012 IEEE RO-MAN: The 21st IEEE International Symposium on Robot and Human Interactive Communication}.\hskip 1em plus 0.5em minus 0.4em\relax IEEE, 2012, pp. 919--924.

\bibitem{mandlekar2018roboturk}
A.~Mandlekar, Y.~Zhu, A.~Garg, J.~Booher, M.~Spero, A.~Tung, J.~Gao, J.~Emmons, A.~Gupta, E.~Orbay, \emph{et~al.}, ``Roboturk: A crowdsourcing platform for robotic skill learning through imitation,'' in \emph{Conference on Robot Learning}.\hskip 1em plus 0.5em minus 0.4em\relax PMLR, 2018, pp. 879--893.

\bibitem{whitney2019comparing}
D.~Whitney, E.~Rosen, E.~Phillips, G.~Konidaris, and S.~Tellex, ``Comparing robot grasping teleoperation across desktop and virtual reality with ros reality,'' in \emph{Robotics Research: The 18th International Symposium ISRR}.\hskip 1em plus 0.5em minus 0.4em\relax Springer, 2019, pp. 335--350.

\bibitem{zhang2018deep}
T.~Zhang, Z.~McCarthy, O.~Jow, D.~Lee, X.~Chen, K.~Goldberg, and P.~Abbeel, ``Deep imitation learning for complex manipulation tasks from virtual reality teleoperation,'' in \emph{2018 IEEE international conference on robotics and automation (ICRA)}.\hskip 1em plus 0.5em minus 0.4em\relax Ieee, 2018, pp. 5628--5635.

\bibitem{lipton2017baxter}
J.~I. Lipton, A.~J. Fay, and D.~Rus, ``Baxter's homunculus: Virtual reality spaces for teleoperation in manufacturing,'' \emph{IEEE Robotics and Automation Letters}, vol.~3, no.~1, pp. 179--186, 2017.

\bibitem{jaegle2021perceiver}
A.~Jaegle, F.~Gimeno, A.~Brock, O.~Vinyals, A.~Zisserman, and J.~Carreira, ``Perceiver: General perception with iterative attention,'' in \emph{International conference on machine learning}.\hskip 1em plus 0.5em minus 0.4em\relax PMLR, 2021, pp. 4651--4664.

\bibitem{chen2023genaug}
Z.~Chen, S.~Kiami, A.~Gupta, and V.~Kumar, ``Genaug: Retargeting behaviors to unseen situations via generative augmentation,'' \emph{arXiv preprint arXiv:2302.06671}, 2023.

\bibitem{mandi2022cacti}
Z.~Mandi, H.~Bharadhwaj, V.~Moens, S.~Song, A.~Rajeswaran, and V.~Kumar, ``Cacti: A framework for scalable multi-task multi-scene visual imitation learning,'' \emph{arXiv preprint arXiv:2212.05711}, 2022.

\bibitem{yu2023scaling}
T.~Yu, T.~Xiao, A.~Stone, J.~Tompson, A.~Brohan, S.~Wang, J.~Singh, C.~Tan, J.~Peralta, B.~Ichter, \emph{et~al.}, ``Scaling robot learning with semantically imagined experience,'' \emph{arXiv preprint arXiv:2302.11550}, 2023.

\bibitem{hirose2023exaug}
N.~Hirose, D.~Shah, A.~Sridhar, and S.~Levine, ``Exaug: Robot-conditioned navigation policies via geometric experience augmentation,'' in \emph{2023 IEEE International Conference on Robotics and Automation (ICRA)}.\hskip 1em plus 0.5em minus 0.4em\relax IEEE, 2023, pp. 4077--4084.

\bibitem{bahl2022human}
S.~Bahl, A.~Gupta, and D.~Pathak, ``Human-to-robot imitation in the wild,'' \emph{arXiv preprint arXiv:2207.09450}, 2022.

\bibitem{chen2024mirage}
L.~Y. Chen, K.~Hari, K.~Dharmarajan, C.~Xu, Q.~Vuong, and K.~Goldberg, ``Mirage: Cross-embodiment zero-shot policy transfer with cross-painting,'' \emph{arXiv preprint arXiv:2402.19249}, 2024.

\bibitem{brooks2023instructpix2pix}
T.~Brooks, A.~Holynski, and A.~A. Efros, ``Instructpix2pix: Learning to follow image editing instructions,'' in \emph{Proceedings of the IEEE/CVF conference on computer vision and pattern recognition}, 2023, pp. 18\,392--18\,402.

\bibitem{rombach2022high}
R.~Rombach, A.~Blattmann, D.~Lorenz, P.~Esser, and B.~Ommer, ``High-resolution image synthesis with latent diffusion models,'' in \emph{Proceedings of the IEEE/CVF conference on computer vision and pattern recognition}, 2022, pp. 10\,684--10\,695.

\bibitem{Dwibedi_2019_CVPR}
D.~Dwibedi, Y.~Aytar, J.~Tompson, P.~Sermanet, and A.~Zisserman, ``Temporal cycle-consistency learning,'' in \emph{The IEEE Conference on Computer Vision and Pattern Recognition (CVPR)}, June 2019.

\bibitem{ravi2024sam}
N.~Ravi, V.~Gabeur, Y.-T. Hu, R.~Hu, C.~Ryali, T.~Ma, H.~Khedr, R.~R{\"a}dle, C.~Rolland, L.~Gustafson, \emph{et~al.}, ``Sam 2: Segment anything in images and videos,'' \emph{arXiv preprint arXiv:2408.00714}, 2024.

\bibitem{yu2023inpaint}
T.~Yu, R.~Feng, R.~Feng, J.~Liu, X.~Jin, W.~Zeng, and Z.~Chen, ``Inpaint anything: Segment anything meets image inpainting,'' \emph{arXiv preprint arXiv:2304.06790}, 2023.

\bibitem{mur2015orb}
R.~Mur-Artal, J.~M.~M. Montiel, and J.~D. Tardos, ``Orb-slam: A versatile and accurate monocular slam system,'' \emph{IEEE transactions on robotics}, vol.~31, no.~5, pp. 1147--1163, 2015.

\end{thebibliography}
}

\end{document}